\documentclass{article}
\usepackage{amsmath,graphicx,tabularx,booktabs,color}
\usepackage[preprint]{spconf}
\usepackage[font={md}, textfont=md]{caption}
\makeatletter 
    \newcommand\fcaption{\def\@captype{figure}\caption}
\makeatother

\copyrightnotice{\copyright\ IEEE 2019}
\toappear{To appear in {\it The 26th IEEE International Conference on Image Processing (ICIP)}}

\title{Multi-Stream Single Shot Spatial-Temporal Action Detection}
\name{Pengfei Zhang, Yu Cao, Benyuan Liu}%\thanks{Thanks to XYZ agency for funding.}
\address{Department of Computer Science, University of Massachusetts Lowell, USA}
\begin{document}
\maketitle
\begin{abstract}
We present a 3D Convolutional Neural Networks (CNNs) based single shot detector for spatial-temporal action detection tasks. Our model includes: ({\romannumeral1}) two short-term appearance and motion streams, with single RGB and optical flow image input separately, in order to capture the spatial and temporal information for the current frame; ({\romannumeral2}) two long-term 3D ConvNet based stream, working on sequences of continuous RGB and optical flow images to capture the context from past frames. Our model achieves strong performance for action detection in video and can be easily integrated into any current two-stream action detection methods. We report a frame-mAP of $71.30\%$  on the challenging UCF101-24 \cite{journals/corr/abs-1212-0402} actions dataset, achieving the state-of-the-art result of the one-stage methods. To the best of our knowledge, our work is the first system that combined 3D CNN and SSD in action detection tasks.
\end{abstract}

\section{Introduction}
\label{sec:intro}
The objective of action detection is to recognize and localize all the human action instances in a given video across both space and time. It is a fundamental task for video understanding and important for practical applications such as video surveillance and human-robot interaction. Action detection is a challenging problem due to two main difficulties: ({\romannumeral1}) it is hard to capture visual representations in the large spatial-temporal search space; ({\romannumeral2})  it is difficult to understand the video fast and accurately, while the detection speed is essential for many application scenarios such as fall and violence detection. %An online algorithm is thus needed to processes each incoming video frame in a timely fashion.

To investigate the spatial-temporal video representation, researchers leverage the hand-crafted features such as dense trajectory \cite{NIPS2014_5353,6751553} and optical flow \cite{NIPS2014_5353,7410719} to build a two-stream network \cite{6909619} to combine the spatial-temporal information together. However, most of these approaches overlook a fundamental issue in action detection, namely, the specific representation of spatial-temporal information for various actions. Many of them only use optical flow, an estimation of motion for each pixel between two images, as the source of temporal information. Conventional approaches estimate optical flows between adjacent frames, which could only represent temporal information of short time periods but lack of long-term information that is important for human action recognition as well. With the successes that 2D CNN has achieved in the field of visual representation on spatial domain, it is natural to extend it to 3D to capture both spatial and temporal information. In action recognition tasks, even though the 3D CNN (I3D \cite{8099985}) achieved the best result so far, the improvement brought by 3D CNN, compared to the hand-crafted features based 2D CNN approaches \cite{8237852}, has not reached its full potential. %5995407,7410734,BMVC2015_177,7410550
% Furthermore, {\color{red}{Furthermore oreFurthe rmoreFur thermo reFurthermoreFurther moreFurtherm oreFurthermoreFurth ermoreFurthermore(这里强调一下hand crafted feature在分类任务中的劣势)}}. 
% ,TSN2016ECCV,hidden_ar_zhu_2017,ren15fasterrcnn,peng:hal-01349107,8237655,8237734,7298676,7410719
In this work, we revisit the role of optical flow and 3D convolution in temporal reasoning for action detection. To explore the contribution of the 4 streams: 2D RGB, 2D optical flow (OF), 3D RGB and 3D OF in action detection, we propose a multi-stream architecture and examine the performance of different stream combinations for various types of actions. We demonstrate that, for the single-stream framework, 3D CNN based model outperforms 2D CNN based model for RGB and optical flow respectively. However, in a two-stream framework, there are different winners of appearance and motion stream for various actions due to the large intra-class variability. As a result, the best frame level mean average precision (mAP) is achieved by the fusion of all four streams, which adapts to a variety of actions.

As for the second challenge on detection speed, although many conventional action detection methods \cite{7410719,peng:hal-01349107,7298676,8578731,ren15fasterrcnn} achieved good results, their two-stage architecture performs region proposal and classification in two steps. While the accuracy is improved, it significantly slows down the detection speed, making it unacceptable for realistic scenarios. To accelerate the detection speed, inspired by \cite{8237655,8237734}, our model adopts the one-stage method, Single Shot MultiBox Detector (SSD) \cite{DBLP:conf/eccv/LiuAESRFB16}, as the detection framework. It merges the two stages into a single network, carries out the localization and classification simultaneously, and thus accelerates the entire process.

The key contributions of this paper include: ({\romannumeral1}) we leverage the single stage object detection architecture SSD to build a time efficient action detector; ({\romannumeral2}) we explore different combinations of 2D and 3D streams for the detection task for a variety of action videos; ({\romannumeral3}) experiment results show that our model outperforms previous one-stage action detection methods on the challenging untrimmed sports video dataset UCF101-24.

\section{Related work}
Our research builds on previous works in two fields:\\
\label{sec:related_work}
\indent\textbf{Spatial-temporal action localization.} Gkioxari and Malik \cite{7298676} applied a two-stream R-CNN based framework to produce frame level detections, and then linked the result to tubes with a dynamic programming method. Weinzaepfel \textit{et al.} \cite{7410719} extracted EdgeBoxes as the action proposals and then used a tracking-by-detection method instead of the linking method. Both Saha \textit{et al.} \cite{ren15fasterrcnn} and Peng \textit{et al.} \cite{peng:hal-01349107} leveraged two-stream Faster R-CNN to do action detection. Singh \textit{et al.} \cite{8237655} applied a single stage detection method SSD to perform online detection. Kalogeiton \textit{et al.} \cite{8237734} extend SSD's anchor boxes to anchor cuboids to perform the temporal-spatial proposal.\\
\indent\textbf{3D CNN.} Ji \textit{et al.} \cite{6165309} and Tran \textit{et al.} \cite{7410867} extended the 2D convolutional kernel to 3 dimensions, and much subsequent studies such as I3D and P3D \cite{8237852} has gained lots of successes in video related tasks. The most recent state-of-the-art result is achieved by Gu \textit{et al.} \cite{8578731} based on I3D and faster-RCNN. Hou \textit{et al.} \cite{hou2017end} designed a C3D version of one-stage action detection method, however, it is an offline algorithm could not do frame level incremental detection.
\section{Model Description}
\label{sec:S3AD}
\label{ssec:network}
\begin{figure}[t]
    \centering
    \includegraphics[width=7.9cm]{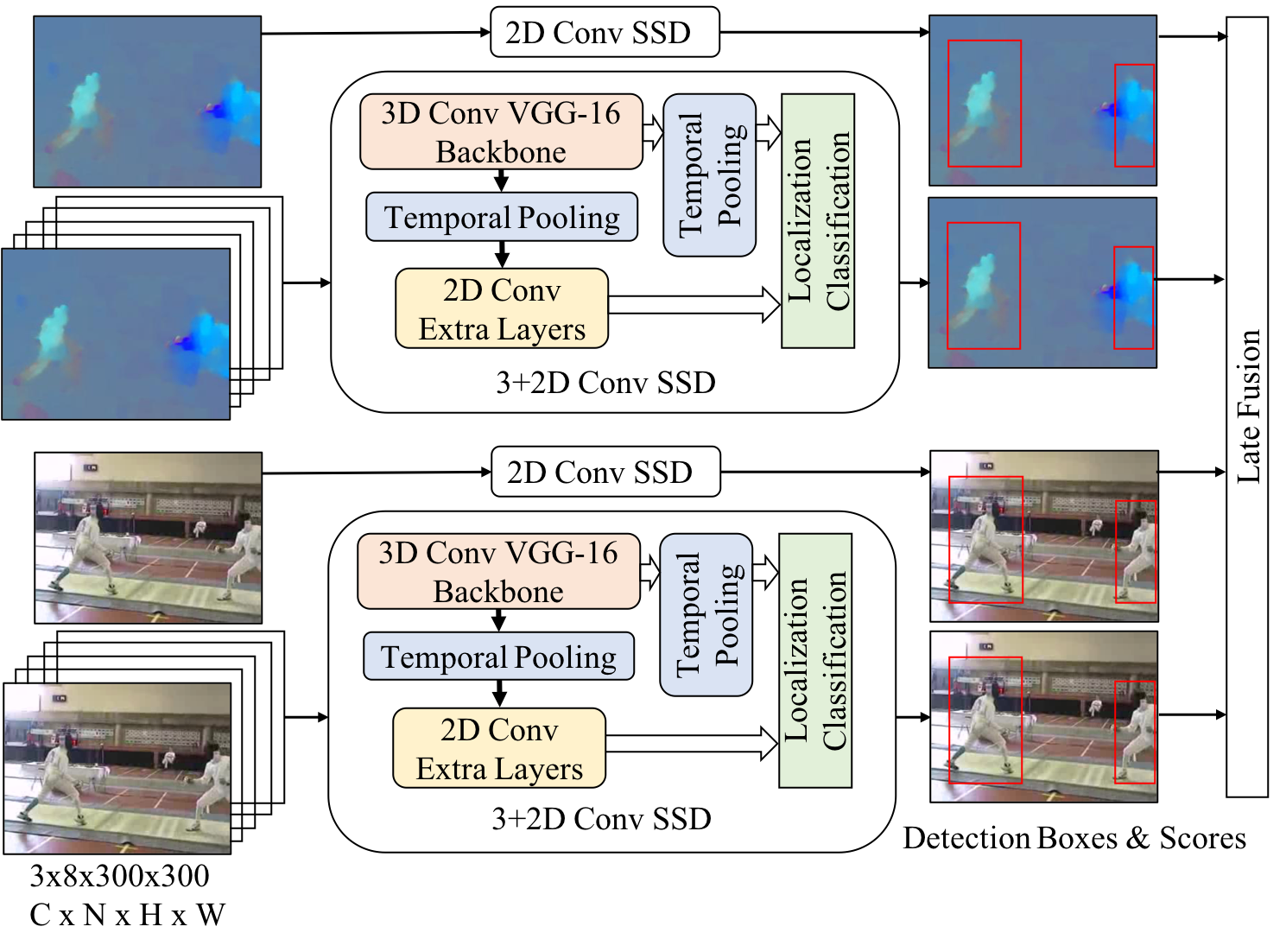}
    \caption{Illustration of the proposed two-stream architecture. 3D SSD takes consecutive video frames as input and extracts both spatial and temporal information.}
    \label{fig:pipeline}
\end{figure}

\textbf{Multi-stream model.} The architecture of our model is illustrated in Fig.\ref{fig:pipeline}. Our model consists of 4 streams: 2D and 3D RGB streams, 2D and 3D optical flow streams. The conventional 2D RGB and optical flow streams are employed to capture the short-term spatial-temporal features, meanwhile, 3D streams are added to learn long-term features. The two 2D streams share the same architecture, but are trained individually and have their own parameters. The same applies to the 3D streams.

For the target action instances at time $t$, the 2D RGB stream's input is current frame $f_t$, while the input of the 2D optical flow stream is extracted from the pair of $\{f_{t-1},f_t\}$ using Brox \textit{et al.}’s \cite{Brox2004} method. The input dimension of both 2D streams is $C\times H\times W$, where $C$, $H$ and $W$ denote the number of channels, height and width of the input frame, respectively. To perform spatial-temporal reasoning with 3D CNN, 3D RGB stream's input is a sequence of continuous $N$ frames $\{f_{t-N+1}...f_t\}$. Similarly, 3D optical flow stream's input is $N$ frames extracted from RGB frame pairs from $\{f_{t-N},f_{t-N+1}\}$ to $\{f_{t-1},f_{t}\}$. The input dimension is $C\times N\times H\times W$. We set $N = 8$ frames in the experiment.\\
\indent\textbf{2D SSD network.} Each of the 2D networks consists of 3 main parts: backbone network, extra convolutional layers and detection heads. The backbone network is truncated VGG-16 and its last two fully connected layers fc6 and fc7 are converted to convolutional layers. Eight extra layers are added to the end of the backbone network to predict default bounding boxes' offsets and their
confidences for actions. Each of the selected layers has a different spatial output dimension, that represents the action instance in different scales. The final predictions are produced by two detection heads: localization head and classification head, synchronously. We use a VGG-16 model pretrained on ImageNet to initialize our model and fine-tune it on the action dataset.\\
% \begin{center}
%     \includegraphics[width=8cm]{ICIP_figure_0.png}
%     \fcaption{Illustration of the proposed two-stream architecture. 3D SSD takes consecutive video frames as input and extracts both spatial and temporal information.}\label{fig:pipeline}
% \end{center}
% \begin{center}
%     \includegraphics[width=8.5cm]{ICIP_figure_1.png}
%     \fcaption{Details of the inflated 3D backbone.}\label{fig:detail}
% \end{center}
\indent\textbf{3D SSD network.} As for the 3D streams, keeping the extra layers and detection heads unchanged, we inflate all the convolutional and pooling layers in backbone network from 2D to 3D, then apply temporal pooling to bridge the gap between 3D and 2D networks. To initialize the network, we repeat the weights of pretrained model's 2D kernels \textit{T} times, where \textit{T} represents the size of the inflated kernel in temporal dimension. In our model, we convert all 3$\times$3 kernels to 3$\times$3$\times$3 kernels, set all layers' temporal padding as $1$ and temporal stride for pooling layers as $2$.\\
\indent\textbf{Temporal pooling.} We connect 3D and 2D layers by the temporal pooling layer. This layer performs mean-pooling along the temporal dimension, transforming the input feature map with dimension $C\times N\times H\times W$ to the output with dimension of $C\times H\times W$.\\
\begin{figure}[htb]
    \centering
    \includegraphics[width=8cm]{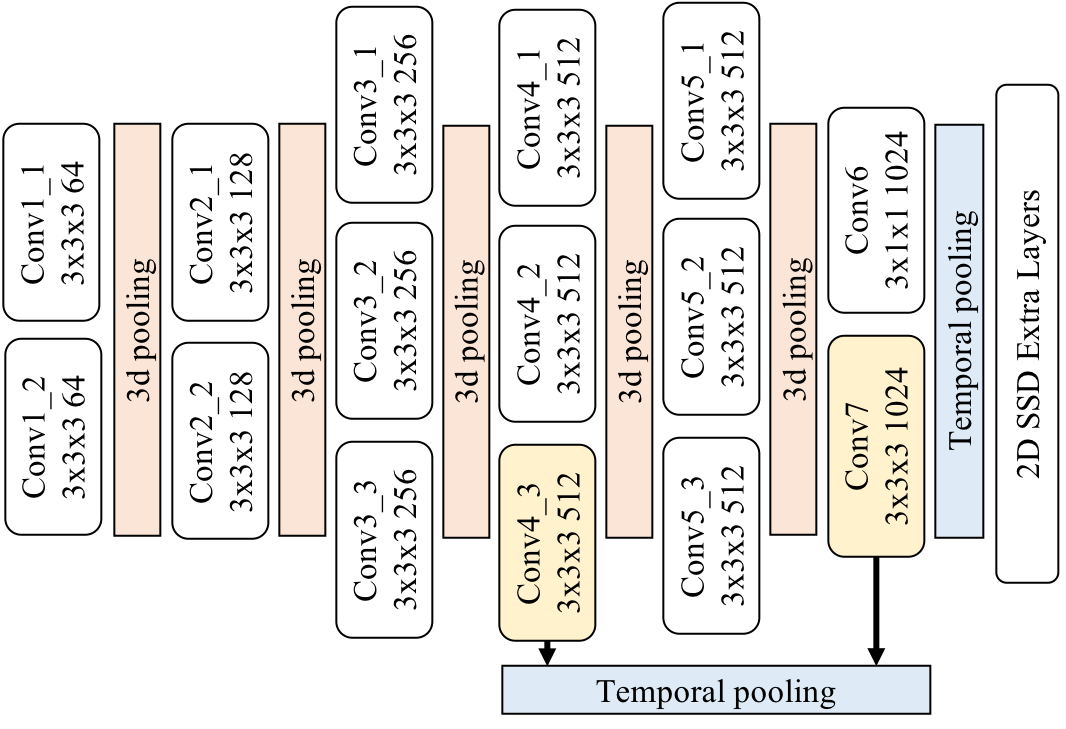}
    \caption{Details of the inflated 3D backbone.}
    \label{fig:detail}
\end{figure}
\begin{figure*}[t]%htb
  \centering
  \centerline{\includegraphics[width=18cm]{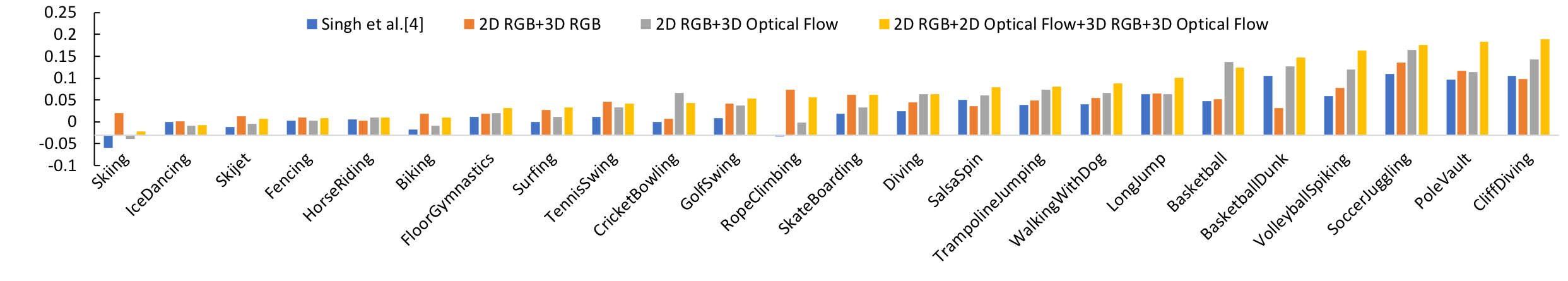}}
  \caption{UCF101-24 frame average precision for each action class compared to 2D RGB with baseline, the value of each class is compute with $AP_{multi-stream} - AP_{2D RGB}$.} \medskip
  \label{fig:all_AP}
\end{figure*}
\indent\textbf{Fusion Method.} We adopt late-fusion \cite{6909619,NIPS2014_5353,7780582} to merge the spatial and temporal information from each stream together. In this step, we first choose one stream as the appearance stream, such as the 2D or 3D RGB stream, keep its bounding boxes regression result, and then set each box's confidence score as the average score of the corresponding boxes from all fused streams. In the rest of paper, we will denote $A+M_1+M_2+...M_n$ as the late fusion of appearance stream $A$ and motion streams from $M_1$ to $M_n$, $n$ denotes the number of motion streams.
\section{Experiments}
\label{sec:experiment}
To evaluate the performance of 3D SSD stream, we examine different stream combinations and their detection accuracy on the UCF101-24 dataset. Singh's 2D SSD real-time framework \cite{8237655} is used as a baseline. We keep their fusion and linking methods unchanged and focus on the performance improvement resulted from 3D SSD.
\subsection{Settings}
\label{ssec:setting}
\indent\textbf{Datasets.} We choose the first split of UCF101-24 dataset to evaluate our model. It contains 24 sport classes in 3,207 untrimmed videos. Each video is annotated with bounding boxes for each action instance at frame level and each frame may contain multiple actors. \\%As in the \cite{8237882,8237655,8237734}
\indent\textbf{Evaluation metrics.} We evaluate the detection accuracy by mean average precision (mAP) for both frame and video levels. At frame level, if the Intersection-over-Union (IoU) between a predicted bounding box and ground truth is greater than a threshold $\alpha$ and this box's action category is classified correctly, we will mark it as an correct detection. As for video level metric, after we connect the frame level detection into tubes, we can evaluate it with the spatial-temporal overlap between the predicted and annotated tubes. As in \cite{7298676,7410719,ren15fasterrcnn}, we present the performance of our model in Table \ref{tab:one_stream}, \ref{tab:two_stream}, and \ref{tab:multi_stream_fmAP}, for frame-mAP with IoU threshold 0.5 and video-mAP with multiple IoU thresholds, $\alpha=0.2, \alpha=0.5, \alpha=0.5:0.95$.
\subsection{Performance}
\label{ssec:performance}
We will first analyse the performance of single streams, then further discuss the contribution of each stream in an ablation study of different two-stream combinations, and show how to get the best result for various of data at last.

\textbf{Single-Stream.} We report the comparison of 2D and 3D streams for RGB and optical flow in Table \ref{tab:one_stream}. The 2D streams adopt the same architecture and experiment setup as in Singh et.al \cite{8237655}. For both frame and video level, each of our 3D streams outperforms the corresponding 2D streams, especially at video level, our 3D RGB network improves the mAP by $5.53\%$ and $4.79\%$ for IoU threshold $\alpha = 0.2$ and $\alpha = 0.5$,  respectively. Similarly, our 3D OF network improves $3.76\%$ and $4.46\%$. The result indicates that the temp†oral information brought in by 3D convolution significantly improves the single-stream model's performance.\\
\begin{table}[!t]
    \small
    \centering
    \begin{tabular}{lcccc}
    \toprule
        Method                         &                   &video-mAP              &               &f.-mAP\\
        IoU                            &0.2                &0.5                    &0.5:0.95       &0.5\\
    \midrule
        2D RGB \cite{8237655}  &69.8	           &40.9		           &18.7           &64.96\\
        2D OF \cite{8237655}   &63.7	           &30.8		           &11.0           &47.26\\ 
    \midrule
        ours-3D RGB                    &\textbf{75.33}     &\textbf{45.69}         &\textbf{19.15} &\textbf{65.10}\\
        ours-3D OF                     &\textbf{67.46}     &\textbf{35.26}         &\textbf{12.51} &\textbf{50.85}  \\
    \bottomrule
    \end{tabular}
    \caption{Comparison of video and frame mAP between 2D and 3D RGB and Optical Flow (OF) streams.}
    \label{tab:one_stream}
\end{table}
% \begin{center}
%     \includegraphics[width=8.5cm]{ICIP_figure_2_tight.png}
%     \fcaption{Comparison between the frame level average precision result of 2D RGB and our 3D RGB(m) fusion and the fusion of 2D RGB and 2D OF \cite{Brox2004, 8237655}, the value of each class is compute as $AP_{2D RGB +3D RGB} - AP_{2D RGB +2D OF}$.}\label{fig:3DRGB}
% \end{center}
\indent\textbf{Two-Stream.} In this section, we answer the following two questions: ({\romannumeral1}) which stream is the best appearance stream? ({\romannumeral2}) which stream is the best motion stream?\\
\indent As for appearance stream, because 3D RGB stream contains both spatial and temporal information, our model can choose either 2D or 3D RGB stream as appearance stream. As shown in Table \ref{tab:two_stream}, the result of 2D RGB $+$ 2D OF outperform that of 3D RGB $+$ 2D OF by $0.75\%$ with the same fusion method. Meanwhile, when the motion stream is 3D OF, the combination with 2D RGB appearance stream outperforms that of 3D RGB by $0.75\%$. This can be explained as the 2D RGB stream contains more accurate spatial information for current frame, while the 3D convolution brings in certain noises from the previous frames.\\
\begin{table}
    \small
    \centering
    \begin{tabular}{lp{0.6cm}<{\centering}p{0.6cm}<{\centering}p{1cm}<{\centering}c}
    \toprule
        Method                                  &\multicolumn{3}{c}{video-mAP}                          & f.-mAP\\
        IoU                                     &0.2                &0.5                &0.5:0.95       & 0.5\\
    \midrule
        2D RGB+2D OF (b)\cite{8237655}  &73.0	            &44.0		        &19.2	        &68.31\\
        2D RGB+2D OF (u)\cite{8237655}  &73.5	            &46.3		        &20.4	        &64.97\\ 
        2D RGB+2D OF (l)\cite{8237655}  &76.43              &45.18              &20.08          &67.81\\ 
    \midrule
        ours-3D RGB+2D OF(l)                         &76.02	            &47.38		        &19.35          &67.06\\
        ours-2D RGB+3D RGB(l)                        &76.18	            &46.52	            &20.94          &68.72\\
        ours-3D RGB+3D OF(l)                         &76.84	            &46.38	            &19.2           &68.82\\
        ours-2D RGB+3D OF(l)                         &\textbf{77.19}	    &\textbf{47.75}		&\textbf{21.11} &\textbf{69.47}\\
    \bottomrule
    \end{tabular}
    \caption{Comparison between different combinations of two-stream fusion. (b) boost fusion, (l) late fusion, (u) union fusion.}
    \label{tab:two_stream}
\end{table}
\indent The candidates for motion stream are: 3D RGB, 2D and 3D optical flows. Comparing the 3D RGB stream with 2D optical flow stream, we find that 2D RGB $+$ 3D RGB performs better than 2D RGB $+$ 2D OF in frame-mAP and video-mAP for IoU threshold $\alpha=0.2$ and $0.5:0.95$. The more detailed frame level average precision analysis for each of the 24 action classes is demonstrated in Fig.\ref{fig:all_AP}. Based on the way how actors and background change with respect to camera, the videos of UCF101-24 can be divided into 3 categories: ({\romannumeral1}) \textbf{active background videos}: videos where the camera moves along with the actors, meanwhile, the background environment changes sharply, for example, rope climbing, skiing and skateboarding. For these 3 classes of videos, the 2D RGB $+$ 3D RGB combination outperform the 2D RGB $+$ 2D OF combination by $10.75\%$, $7.89\%$ and $4.35\%$, respectively. The poor performance of 2D optical flow stream is caused by the noises produced by the fast changing background. ({\romannumeral2}) \textbf{fixed background videos}: videos where the camera is fixed, the background does not change much and the actors move quickly in short time frame, such as Salsa Spin, Cliff Diving and Basketball Dunk. Because optical flow contains more accurate short-term temporal information than 3D RGB, the performance of 2D RGB $+$ 2D OF is better than that of 2D RGB $+$ 3D RGB. ({\romannumeral3}) For other videos that contain more complex circumstance, 3D RGB stream's contribution is similar to or slightly better than optical flow.\\
\begin{center}
    \includegraphics[width=6cm]{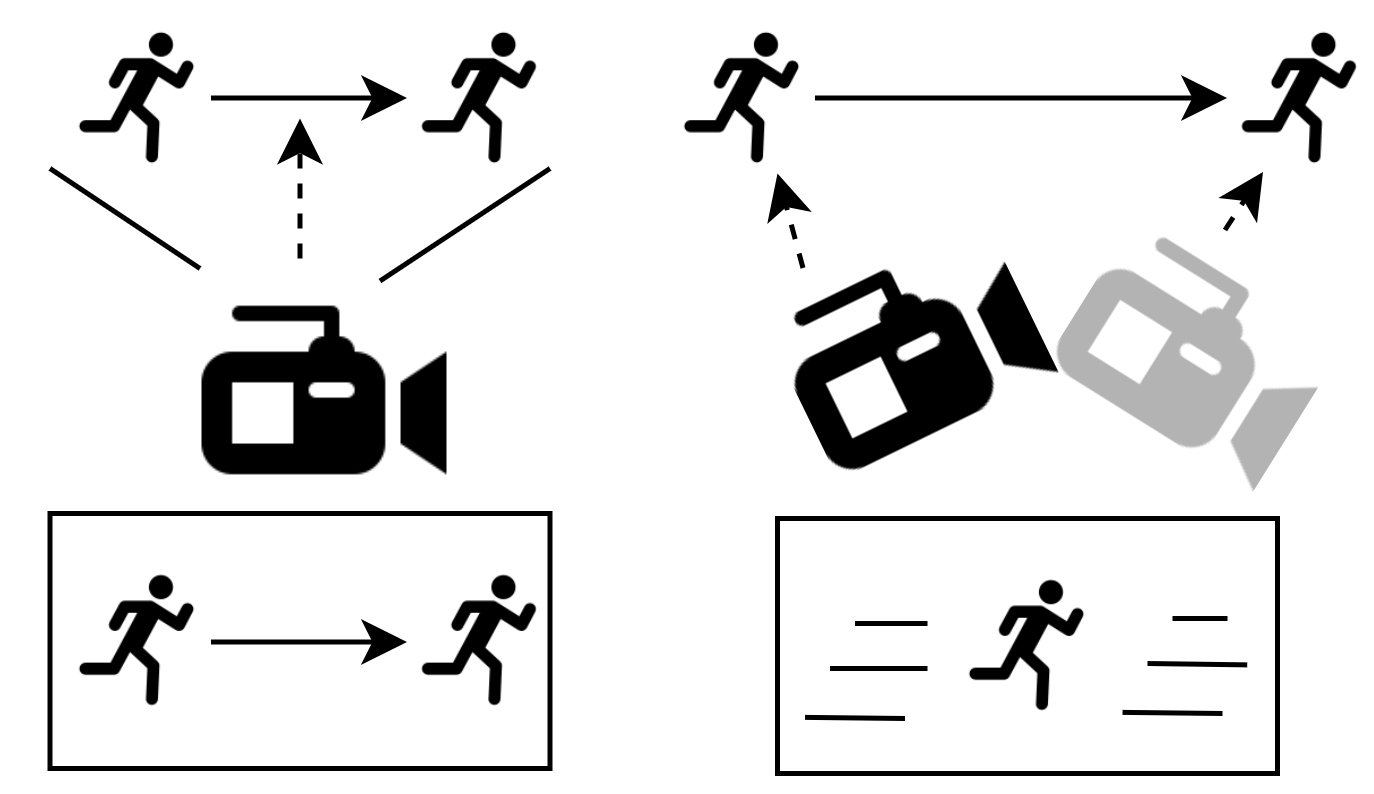}
    \fcaption{Fixed background and Active background videos.}\label{fig:video}
\end{center}
\indent While 2D OF and 3D RGB outperforms each other in different scenarios, the best performance of all two-stream combinations is achieved by 2D RGB $+$ 3D OF. It improve the frame-mAP of 2D RGB $+$ 2D OF by $1.66\%$, and the result of 2D RGB $+$ 3D RGB by $0.75\%$, which means 3D optical flow is the best choice of motion stream in two-stream framework. However, as shown in Fig.\ref{fig:all_AP}, we can observe that the 3D optical flow still inherit the drawback of 2D optical flow, resulting in poor performance for active background videos.\\
\indent\textbf{Multi-Stream.} We present the fusion results of three-stream and four-stream models in Table \ref{tab:multi_stream_fmAP}. Compared to Singh \textit{et al.} \cite{8237655}'s two-stream model, our three-stream model (2D RGB $+$ 3D RGB $+$ 2D OF) obtains $2.23\%$ improvement for the frame-mAP with the 3D RGB stream integrated, and $3.49\%$ improvement with the fusion of all four streams. To the best of our knowledge, our model outperforms all the one-stage methods with better action localization and classification accuracy. In practice, we also need to consider the time consumption to prepare a stack of optical flows, which is important for developing an online real-time system. For different kinds of action videos and applications, our model is flexible to be reorganized or integrated into other models to meet the requirements.
% \indent \\
\begin{table}
    \small
    \centering
    \begin{tabular}{lc}
    \toprule
        Method     &frame-mAP@0.5\\
    \midrule
        % Weinzaepfel \textit{et al.} \cite{7410719} &35.84\\
        % Peng \textit{et al.} \cite{peng:hal-01349107}                    &65.73\\
        (SSD) Kalogeiton \textit{et al.} \cite{8237734}                &67.10\\
        Hou \textit{et al.} \cite{8237882}                           &67.3\\
        (SSD) Singh \textit{et al.} \cite{8237655}                &67.81\\
        \midrule
        ours-2D RGB+3D RGB+2D OF        &70.04\\ 
        ours-2D RGB+3D RGB+3D OF        &71.10\\ 
        ours-3D RGB+3D OF+2D RGB+2D OF        &71.28\\ 
        ours-2D RGB+2D OF+3D RGB+3D OF        &\textbf{71.30}\\ 
        \bottomrule
    \end{tabular}
    \caption{Comparison of frame-mAP to the state-of-the-art on UCF101-24 dataset in split1.}
    \label{tab:multi_stream_fmAP}
\end{table}
\section{Conclusions and future plans}
\label{sec:conclusion}
%这里a写成an了没事吧？？
This paper introduced a multi-stream action detector which achieves state-of-the-art results of the one-stage methods on UCF101-24 dataset. We present an empirical study of the properties of the combinations of 2D RGB, 2D OF, 3D RGB and 3D OF streams. Based on the results of those experiments, the following conclusions could be obtained: ({\romannumeral1}) 2D RGB stream is a better choice for appearance stream comparing to other streams; ({\romannumeral2}) for active background videos, 3D RGB motion stream can tolerate more environmental noises; ({\romannumeral3}) optical flow, especially the 3D stream, performs well for videos that have fixed background and significant short-term action instances. Future work will be devoted to two directions: ({\romannumeral1}) optimize the framework with other one-stage methods, such as YOLO \cite{DBLP:conf/cvpr/RedmonF17} series. ({\romannumeral2}) Improve the temporal convolutional module with more lightweight 3D kernels to accelerate the whole forward process.%Our proposed model outperforms previous single-stage approaches at both the video and frame level on UCF101-24 dataset.
\label{sec:ref}
\bibliographystyle{IEEEbib}
\bibliography{strings,refs}

\end{document}